\documentclass{article}

\usepackage[nonatbib, final]{ARLET_2024}
\usepackage{natbib}
\setcitestyle{authoryear}

\usepackage[utf8]{inputenc} %
\usepackage[T1]{fontenc}    %
\usepackage[backref=page]{hyperref}       %
\usepackage{url}            %
\usepackage{booktabs}       %
\usepackage{amsfonts}       %
\usepackage{nicefrac}       %
\usepackage{microtype}      %
\usepackage{xcolor}         %

\usepackage{graphicx}           %
\usepackage{caption}
\usepackage{subcaption}         %
\usepackage{float}
\definecolor{bluecite}{HTML}{0875b7}
\hypersetup{
  colorlinks=true,
  citecolor=bluecite,
  linkcolor=magenta
}
\usepackage{amsmath}
\usepackage{amssymb}
\usepackage[ruled]{algorithm2e}
\usepackage{bm}

\title{Coordination Failure in Cooperative Offline MARL}

\author{%
  Callum Rhys Tilbury$^1$\thanks{Equal contribution.}
  \And
  Claude Formanek$^{1,2*}$\thanks{Corresponding author: \texttt{c.formanek@instadeep.com}}
  \And
  Louise Beyers$^1$
  \AND
  Jonathan Shock$^{2,3}$
  \And
  Arnu Pretorius$^{1}$\\
  \AND
  \vspace{-1em}\\
  $^{1}$InstaDeep \\
  $^{2}$University of Cape Town \\
  $^{3}$The INRS, Montreal
}

\begin{document}

\maketitle

\renewcommand*{\thefootnote}{\fnsymbol{footnote}}
\begin{abstract}
Offline multi-agent reinforcement learning (MARL) leverages static datasets of experience to learn optimal multi-agent control. However, learning from static data presents several unique challenges to overcome. In this paper, we focus on coordination failure and investigate the role of joint actions in multi-agent policy gradients with offline data, focusing on a common setting we refer to as the `Best Response Under Data' (BRUD) approach. By using two-player polynomial games as an analytical tool, we demonstrate a simple yet overlooked failure mode of BRUD-based algorithms, which can lead to catastrophic coordination failure in the offline setting. Building on these insights, we propose an approach to mitigate such failure, by prioritising samples from the dataset based on joint-action similarity during policy learning and demonstrate its effectiveness in detailed experiments. More generally, however, we argue that \emph{prioritised} dataset sampling is a promising area for innovation in offline MARL that can be combined with other effective approaches such as critic and policy regularisation. Importantly, our work shows how insights drawn from simplified, tractable games can lead to useful, theoretically grounded insights that transfer to more complex contexts. A core dimension of offering is an interactive notebook, from which almost all of our results can be reproduced, in a browser.\footnote[3]{\href{https://tinyurl.com/pjap-polygames}{https://tinyurl.com/pjap-polygames}}
\end{abstract}

\section{Introduction} \label{sec:introduction}

\begin{figure}[t]
    \centering
    \includegraphics[width=\linewidth]{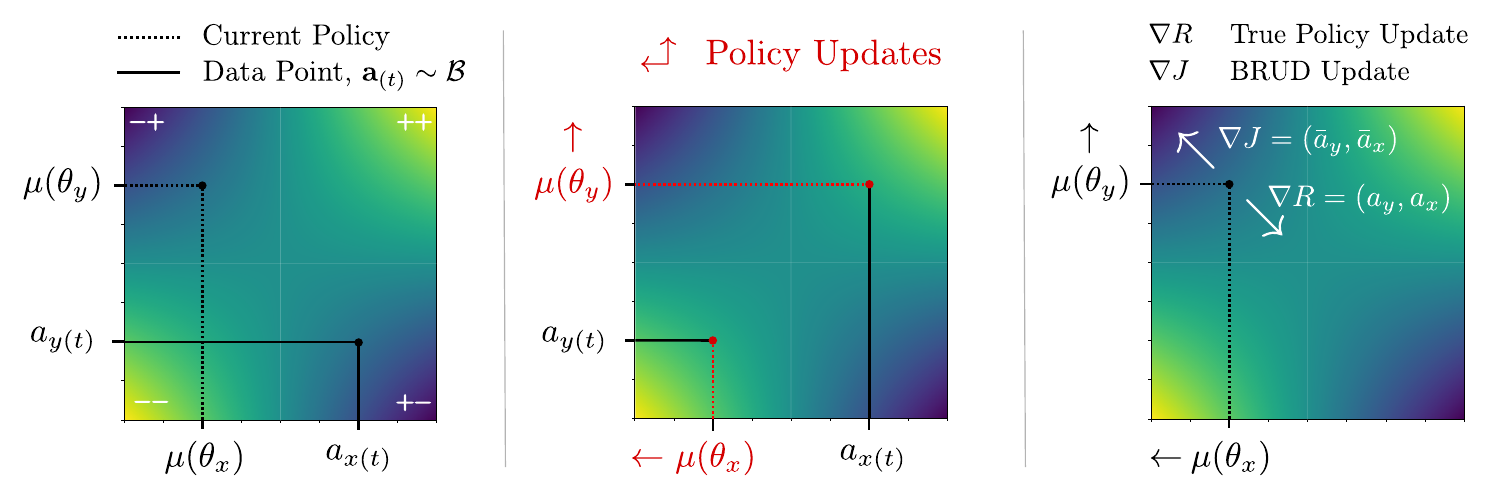}
    \caption{Illustration of catastrophic miscoordination when agents each learn based on a best response to the data of other agent actions (BRUD). We consider using a datapoint $\mathbf{a}_{(t)}$, in a simple game where the reward is given by the product of each agent's action, $R(a_x, a_y)=a_x a_y$. The best response of agent~$x$, in response to the other agent's negative data point, $a_{y(t)} < 0$, is to make its own policy $\mu(\theta_x)$ more negative. Similarly, agent $y$ updates $\mu(\theta_y)$ to be more positive, in response to the other agent's positive data point, $a_{x(t)}>0$. Alas, the BRUD step moves the joint policy in the opposite direction of optimal increase.}
    \label{fig:best-response-illustration}
\end{figure}

Offline reinforcement learning (RL) is a promising paradigm for making real-world applications of RL possible. While some compelling progress is being made, particularly in the single-agent setting~\citep{prudencio2023offlinerl}, large obstacles remain. In this paper, we focus on a problem unique to the multi-agent setting: learning \emph{coordination} from static data~\citep{barde2023modelbased}. Whereas in online multi-agent learning, a speculated failure in coordination can be tested and corrected, such feedback does not exist in the offline case. Instead, the agents are constrained to solely using static data to learn how to best act together. Typically, agents optimise their own actions towards a best response to the actions taken by other agents in the dataset, we refer to this common approach as `Best Response Under Data' (BRUD). This approach has various benefits in the offline setting, but is highly susceptible to miscoordination. This is clearly illustrated in Figure~\ref{fig:best-response-illustration}, using a simple two-player game where agents choose a continuous real number, and the collective reward is the product of the two actions chosen.

In this work, we use simple two-player polynomial games as an analytical tool for better understanding offline coordination, in an interpretable and accessible way. In doing so, we isolate the problem with a BRUD-style update in offline MARL, demonstrating clear modes of coordination failure. Then, building on our insights, we propose a class of offline sampling methods, broadly called Proximal Joint Action Prioritisation (PJAP) that help alleviate problems in coordination that stem from offline learning. We demonstrate the effectiveness of PJAP in detailed experiments. However, we see this work more as exploratory in nature, more generally highlighting prioritised sampling methods as a fruitful area of future investigation alongside approaches such as critic and policy regularisation for offline learning.

\section{Foundations} \label{sec:background}
\subsection{Multi-Agent Reinforcement Learning}

We consider the canonical Dec-POMDP setting for MARL where the goal is to find a joint policy $(\pi_1, \dots, \pi_n) \equiv \bm{\pi}$ such that the return of each agent $i$, following $\pi_i$, is maximised with respect to the other agents’ policies, $\pi_{-i} \equiv (\pi \backslash \pi_i)$. That is, we aim to find $\pi$ such that $\forall i: \pi_i \in {\arg\max}_{\hat{\pi}_i} \mathbb{E}\left[G \mid \hat{\pi}_i, \pi_{-i}\right]$, where $G$ is the return. We assume that each policy is parameterised by $\theta_i$. A popular approach to learning such policies is \textit{Centralised Training with Decentralised Execution} (CTDE), where training leverages privileged information from all agents, yet the policies are only conditioned on their local observations, $\pi_i(o_i; \theta_i)$, enabling decentralisation at inference time. 

Because of our focus on the offline setting, we narrow our scope to off-policy algorithms, where we learn using batches of data taken from a replay buffer, $\mathcal{B}$. Specifically, we study multi-agent actor-critic approaches which have a policy objective of the form,
\begin{equation}
    \label{eqn:policy-objective}
    J(\bm{\pi}) = \mathbb{E}_{\mathbf{a}\sim\bm{\pi}}\left[Q(\mathbf{o}, \mathbf{a}) + \alpha\mathcal{R}\right]
\end{equation}
where $Q(\mathbf{o}, \mathbf{a})$ is the joint critic, and $\mathcal{R}$ is some policy regularisation term, controlled by $\alpha\in\mathbb{R}$. This policy objective is a key component of many popular CTDE algorithms in MARL. For example, by using stochastic policies and entropy regularisation, $\mathcal{R}=\mathcal{H}(\bm{\pi})$, we recover the policy objective of multi-agent soft-actor critic~\citep{Pu2021DecomposedSA}. With deterministic policies, $\pi_i(a_i| o_i) = \mu_i(o_i)$, and setting $\alpha=0$, we recover the policy objective of MADDPG~\citep{lowe2017maddpg}, and so on. Importantly, this policy update forms part of several leading \emph{offline} MARL algorithms, such as CFCQL~\citep{cfcql} and the CTDE form of OMAR~\citep{omar}.

\subsection{Joint Action Formulation}%
In the policy objective in Equation~\ref{eqn:policy-objective}, the training is centralised by conditioning the critic on the joint observation, $\mathbf{o}$, and the joint action, $\mathbf{a}$. The joint observation can be formed as a simple concatenation of agent observations in the sampled data, $\mathbf{o} = (o_{1}, ..., o_{n})$. However, forming the joint action is more complex. For the policy learning of agent $i$, we consider $a_i \sim \pi(\cdot | o_i; \theta_{i})$, but we have several choices for the other agent actions, $a_{-i}$, in the update. For example, we could simply use the other agents' policies directly, $a_{-i} \sim \pi(\cdot | o_{-i}; \theta_{-i})$. However, this approach has been shown to work poorly in offline settings~\citep{cfcql}, likely because we are decoupling policy learning from the dataset. Instead, the prevailing approach to forming the joint action in CTDE methods, both online and offline, is to simply use the samples taken from the buffer or dataset for the other agent actions. That is,
\begin{equation}
\label{eqn:joint-action}
\mathbf{a}_i = ( a_i \sim \pi (\cdot | o_i; \theta_i), \; a_{-i} \sim \mathcal{B} )
\end{equation}

We call this the \emph{Best Response Under Data} (BRUD) approach to policy learning. Though it benefits from staying directly coupled to the dataset in an offline context, it leads to coordination problems, which we will demonstrate shortly.

\subsection{Polynomial Games}
For our exposition, we study two-player polynomial games~\citep{dresher1950polynomial, zhong2024heterogeneous}, as a differentiable, continuous generalisation of discrete matrix games---which have been a common tool for understanding multi-agent algorithms~\citep{rashid2020monotonic, papoudakisBenchmarkingMultiAgentDeep2021}. These games are atemporal and stateless,
comprising two agents, $x$ and $y$, each able to take continuous actions. We denote the respective actions taken as $a_x, a_y\in\mathbb{R}$. The shared reward given to the agents is defined by some polynomial function, $R(a_x, a_y)~=~\sum_{i=0}^{m}~\sum_{j=0}^{n}~c_{ij} a_x^i a_y^j$. Because there is no state, and thus no observations, the notion of maximising the joint Q-function, $Q(\bm{o}, \mathbf{a} = \{a_x, a_y\})$ is equivalent to maximising the reward function $R(a_x,a_y)$ directly. We assume perfect knowledge of the reward function in the game.

\section{Coordination Failure in Offline MARL}
We now study coordination failure in offline MARL due to the BRUD approach in policy learning, using tractable and informative polynomial games. We use MADDPG~\citep{lowe2017maddpg}, which has a BRUD-style policy update for agent $i$,
\begin{align}
    \label{eq:maddpg_update}
    \nabla_{\theta_i} J = \mathbb{E}_{(\bm{o}, \bm{a}) \sim \mathcal{B}} \left [\nabla_{\theta_i} \mu(o_{i}; \theta_i) \cdot \nabla_{\tilde{a}_i} Q(\bm{o}, \tilde{a}_i, \bm{a}_{-i}) |_{\tilde{a}_{i} = \mu (o_{i}; \theta_i)}\right ]
\end{align}
where $\mu(o_i; \theta_i)$ is a deterministic policy, and $\mathcal{B}$ is a replay buffer or dataset.

Recall that the polynomial game setting is stateless, and comprises just two agents, taking actions $a_x$ and $a_y$. For simplicity, let the policy for each agent be a single linear unit, $\mu(\theta_x)=\theta_x$ and $\mu(\theta_y)=\theta_y$ (i.e. the policy parameter directly defines the action). We can thus simplify the policy update such that for agent $x$,
\begin{align}
    \nabla_{\theta_x} J & = \mathbb{E}_{\bm{a} \sim \mathcal{B}} \left [\nabla_{\theta_x}  \mu (\theta_x) \cdot \nabla_{\tilde{a}_x} R(\tilde{a}_x, a_y) |_{\tilde{a}_{x} = \mu (\theta_x)} \right ] = \mathbb{E}_{a_y \sim \mathcal{B}} \left [ \nabla_{\tilde{a}_x} R(\tilde{a}_x, a_y) |_{\tilde{a}_{x} = \theta_x} \right ]
    \label{eqn:polygame-maddpg-update}
\end{align}
and similarly for agent $y$, we have $\nabla_{\theta_y} J = \mathbb{E}_{a_x \sim \mathcal{B}} \left [ \nabla_{\tilde{a}_y} R(a_x, \tilde{a}_y) |_{\tilde{a}_y = \theta_y} \right ]$. Therefore, each component in the gradient of the objective is simply the partial derivative of the agent's reward with respect to the agent's chosen action, in expectation over the actions of the other agent from the replay buffer or the dataset. This equation captures the essence of the policy update of BRUD methods.

To understand the ramifications of forming the joint action in this way, we first study the simple polynomial $R(a_x,a_y)=a_x a_y$, which we call the \emph{sign-agreement} game. The true gradient field of this surface is $\nabla R=(a_y,a_x)$, whereas the objective in the MADDPG update becomes $\nabla J = (\mathbb{E}_{a_y\sim\mathcal{B}}[a_y], \mathbb{E}_{a_x\sim\mathcal{B}}[a_x]) = ( \bar{a}_y, \bar{a}_x)$, where $\bar{a}_x$ and $\bar{a}_y$ are the sample means of the respective actions in the data taken from $\mathcal{B}$. We consider the impact of the difference between $\nabla R$ and $\nabla J$ for this game. Whereas the former is a function of the current policy, always correctly pointing to the optimal direction of policy improvement, the latter is a unidirectional vector field defined solely by the sampled data. As a result, it becomes possible for catastrophic miscoordination in the joint-policy update, as illustrated in Figure~\ref{fig:best-response-illustration}. In this example, the best way to update $\theta_x$, in response to a negative data point, $a_{y(t)} < 0$, is to make $\theta_x$ more negative. Simultaneously, the best way to update $\theta_y$, in response to a positive data point, $a_{x(t)}>0$, is to make $\theta_y$ more positive. Alas, this joint update step actually moves the joint policy into the $-+$ region (top left), yielding a lower reward---the exact opposite of our intention.

\subsection{Connections to Off-Policy Learning}

Importantly, the possibility of miscoordination as demonstrated here is present under any BRUD-style update---regardless of whether we are learning online from a dynamic buffer, or offline from a static dataset. Consider, though, how the impact of the policy gradient update from Equation~\ref{eqn:polygame-maddpg-update} changes as we move from learning online to offline. We can illustrate this shift by studying the size of the replay buffer, $\mathcal{B}$. Recall that introducing a replay buffer improves sample efficiency and stabilises training in deep reinforcement learning~\citep{dqn2013}.
A useful way to understand the buffer is the relationship between its size and the degree to which learning is off-policy. Suppose data of size $b$ is sampled from the buffer to update the agents; then with a buffer size of $b$, agents are using experience immediately after witnessing it---which is exactly on-policy, akin to an approach like REINFORCE~\citep{williamsSimpleStatisticalGradientfollowing}. Naturally, then, as the buffer size increases---where data is replaced less and less frequently---the algorithm becomes increasingly off-policy. In the limit, where the buffer size is infinite and data is never replaced, the setting becomes akin to offline learning, albeit with fresh data still being added.

\begin{figure}[h]
    \centering
    \includegraphics[width=0.8\linewidth]{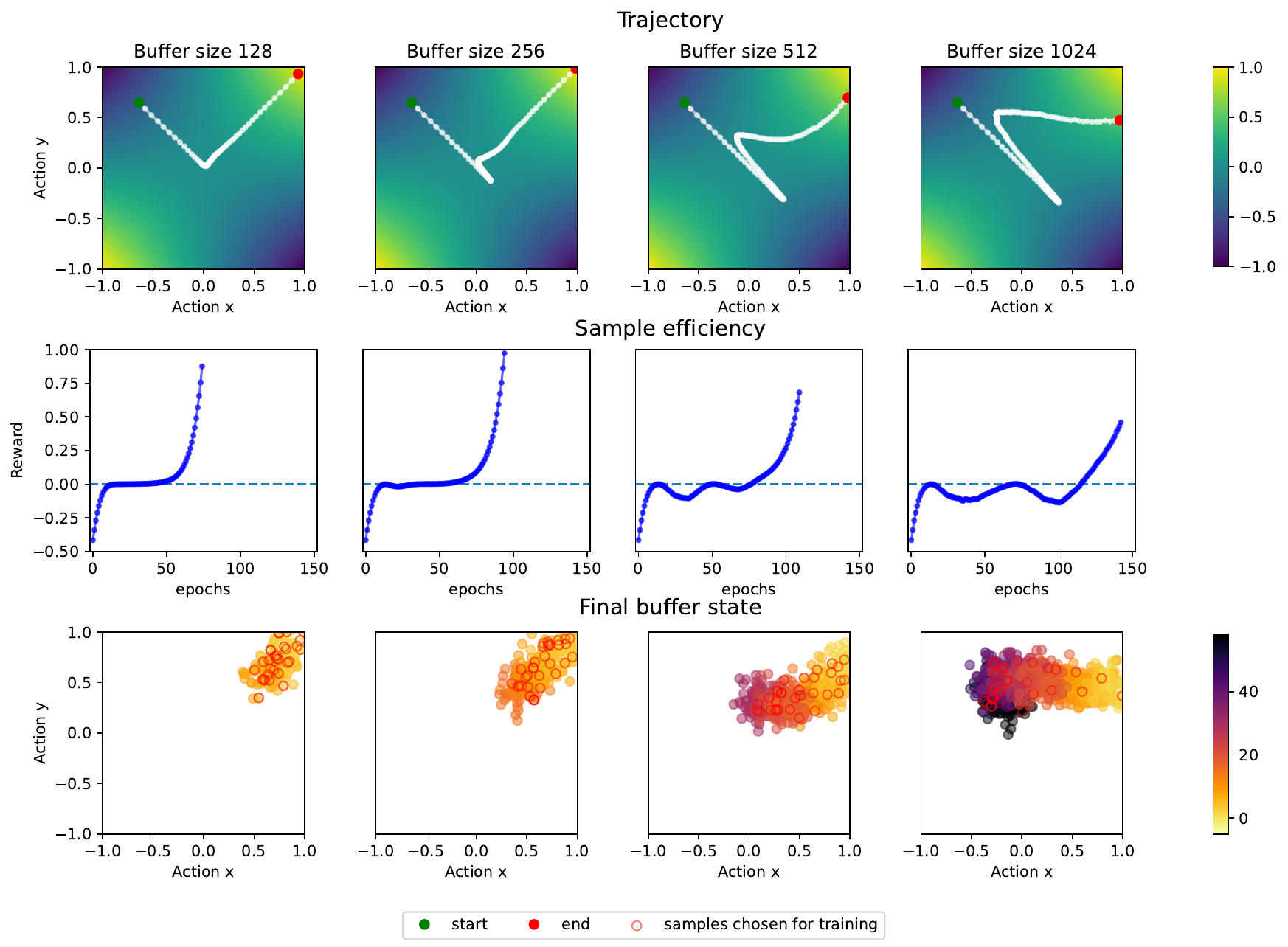}
    \caption{Demonstrating the impact of replay buffer size, as a proxy for off-policyness, on the policy learning with online MADDPG. We show the learning trajectory in policy space (top), the learning as the reward over time (middle), and the state of the replay buffer in the final training update (bottom). We see that increasing the buffer size leads to less optimal trajectories being learnt, due to the presence of the stale data in the replay buffer. With the BRUD update, we can see that it is important for the sampled joint action to remain fairly close to the current joint policy, to avoid miscoordination.}
    \label{fig:off-policyness}
\end{figure}

Figure~\ref{fig:off-policyness} shows the impact of increasing the buffer size in the sign-agreement game when training online with MADDPG. With the smallest buffer size, the policy moves along the optimal trajectory, first towards the saddle at $(0,0)$ and then towards the high-reward region of $++$. The BRUD approach works well here, since the sampled joint action is likely to be relatively close to the current policy, mitigating challenges of miscoordination. Yet as the max buffer size grows (i.e. the algorithm becomes more off-policy), the typical sampled joint-action is further away from the current policy, as visualised in the plots of the buffer state at the end of training. As a result, we see how the learnt policy becomes increasingly sub-optimal, due to miscoordination problems discussed previously.

There is thus a tension between the efficiency and stability introduced by a replay buffer, and the degree to which it is prone to miscoordination. However, we notice that with online MADDPG, in each case, we still find the optimal, high-reward regions eventually. That is, some miscoordination remains possible during training, but through exploration and adding fresh data to the buffer, online learning can nonetheless recover good performance.

\begin{figure}[ht]
    \centering
    \begin{subfigure}[p]{0.31\textwidth}
        \centering
        \includegraphics[width=0.7\linewidth]{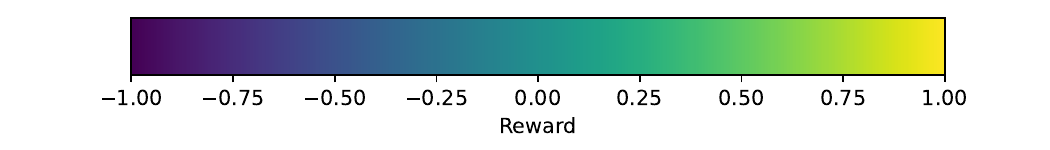}
    \end{subfigure}
    \hfill
    \begin{subfigure}[p]{0.31\textwidth}
        \centering
        \includegraphics[width=0.75\linewidth]{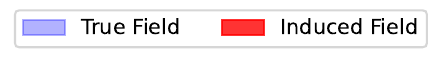}
    \end{subfigure}
    \hfill
    \begin{subfigure}[p]{0.31\textwidth}
        \centering
        \includegraphics[width=0.8\linewidth]{figures/polygames/polygame-legend.pdf}
    \end{subfigure}\vspace{-1em}
        \begin{subfigure}[t]{0.31\textwidth}
            \centering
            \includegraphics[width=0.8\linewidth]{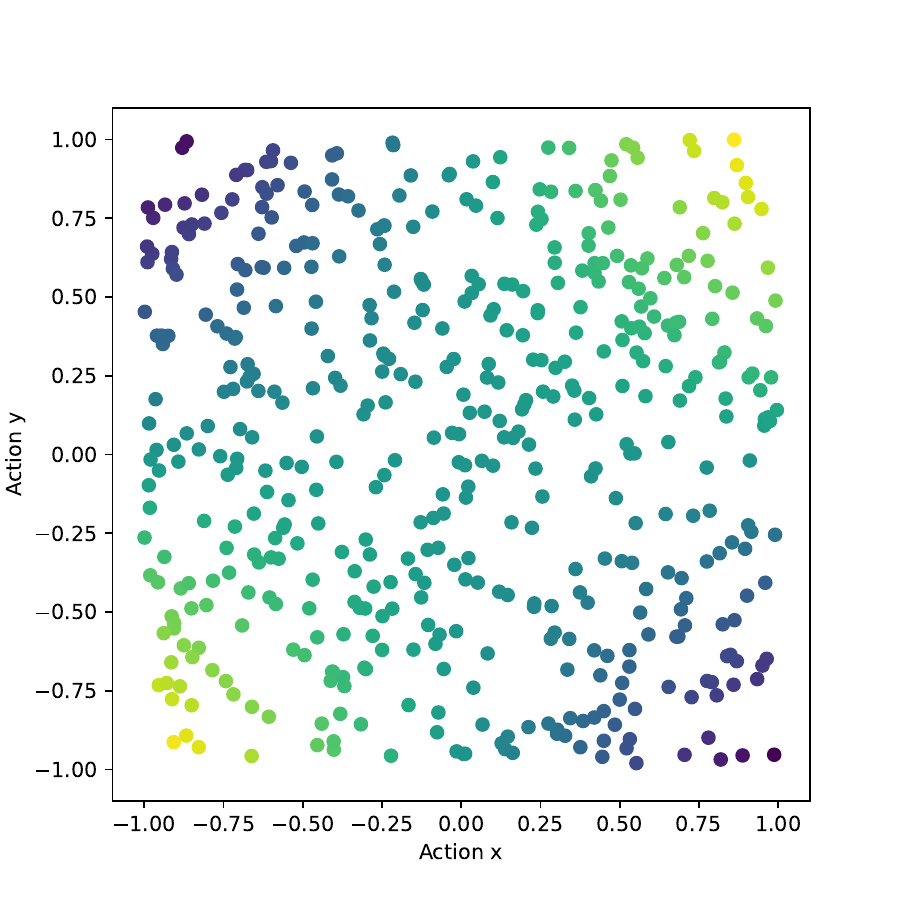}
            \caption{Dataset, $\bar{\mathbf{a}}=(-0.02, 0.04)$}
            \label{fig:offline-polygame-dataset}
        \end{subfigure}
        \hfill
        \begin{subfigure}[t]{0.31\textwidth}
            \centering
            \includegraphics[width=0.8\linewidth]{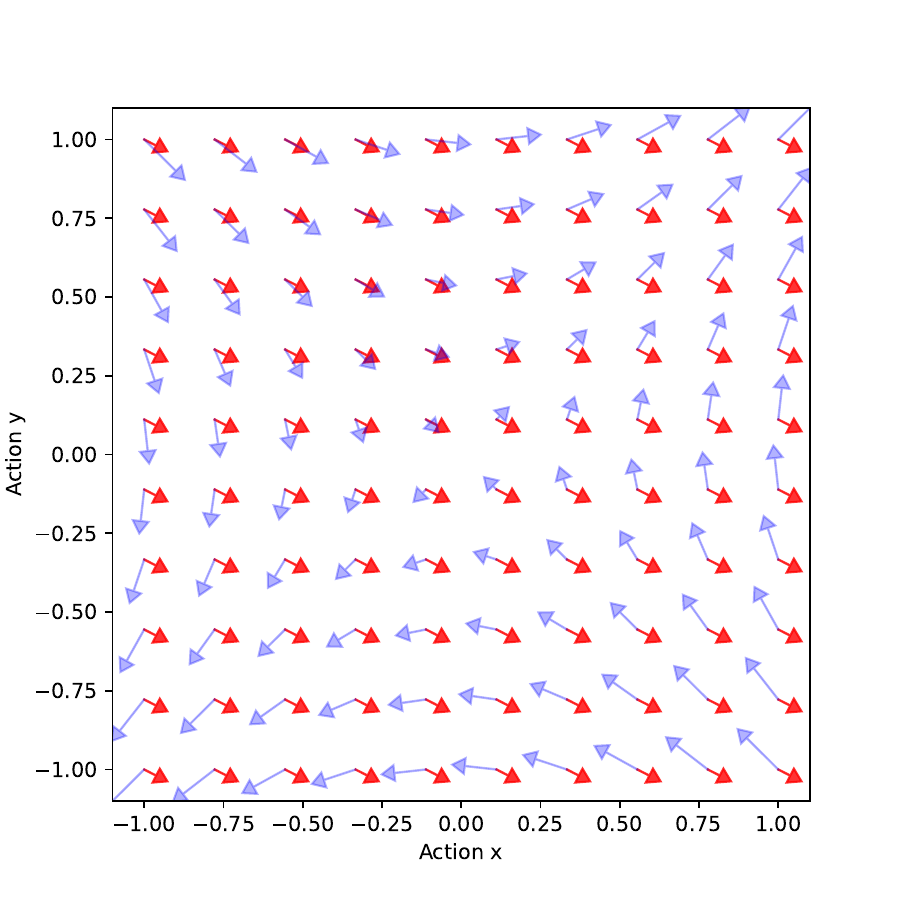}
            \caption{{\textcolor{blue}{$\nabla R$}}, {\textcolor{red}{$\nabla J$}}}
            \label{fig:offline-polygame-vector-field}
        \end{subfigure}
        \hfill
        \begin{subfigure}[t]{0.31\textwidth}
            \centering
            \includegraphics[width=0.7\linewidth]{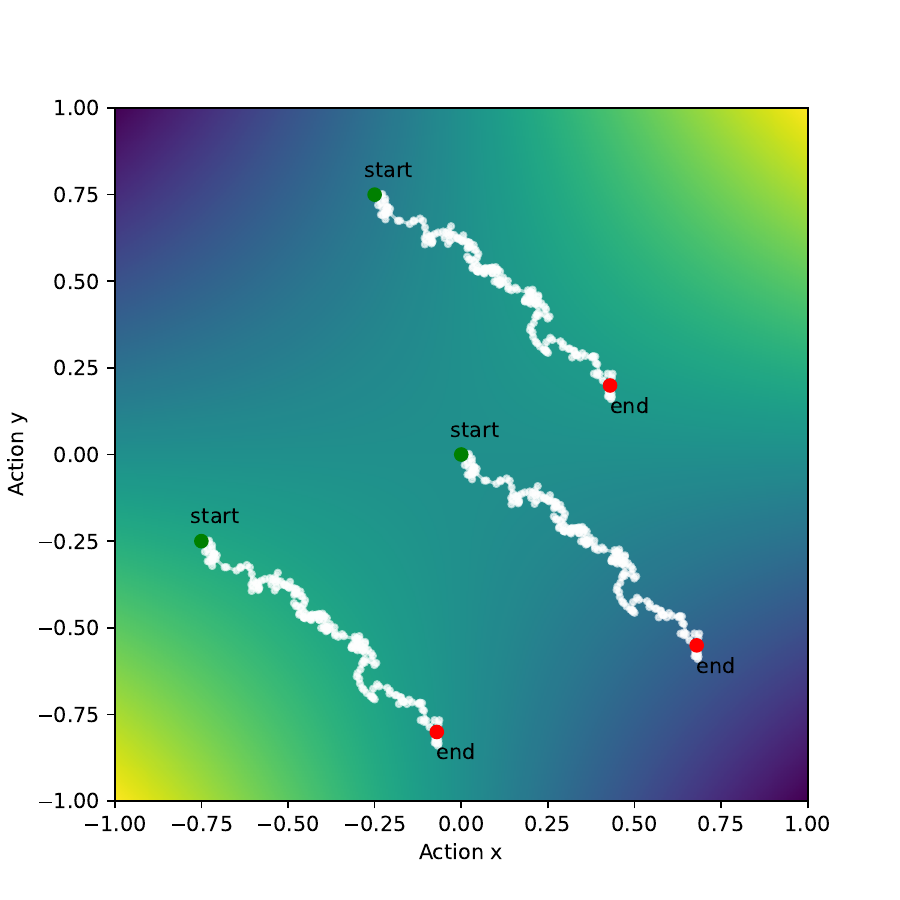}
            \caption{Policy learning trajectories}
            \label{fig:offline-polygame-traj}
        \end{subfigure}
        \caption{The results of using a uniform dataset $\mathcal{B}$ for offline MADDPG policy learning, in the sign-agreement game, $R(a_x, a_y) = a_xa_y$. We see that the net direction of policy learning is predetermined by the mean of the dataset, due to the BRUD approach, regardless of the policy initialisation.}
        \label{fig:offline-uniform-data}
    \label{fig:offline-example}
\end{figure}

Offline learning does not experience the same success. Consider Figure~\ref{fig:offline-example}, which shows an example of learning from a static dataset, sampled uniformly as $\mathcal{B}\sim U(-1,1)$. The data itself, depicted in Figure~\ref{fig:offline-polygame-dataset}, has a small bias, $(\bar{a}_x, \bar{a}_y) = (-0.02, 0.04)$. Recall that in this sign-agreement game, the BRUD update induces a unidirectional vector field $\nabla J = (\bar{a}_y, \bar{a}_x)$, in contrast to the true vector field $\nabla R = (a_x, a_y)$. These two fields are visualised in Figure~\ref{fig:offline-polygame-vector-field}, with all vectors in $\nabla J$ pointing to the bottom right, since $\bar{a}_y > 0$ and $\bar{a}_x < 0$. The resulting trajectories of offline learning with MADDPG, starting from three separate policy initialisations, are shown in Figure~\ref{fig:offline-polygame-traj}. Notice how the nature of the static dataset used for training completely determines the direction of policy update---which, in this case, is in a completely incorrect direction, towards a low-reward region.

\subsection{Growing Risk of Miscoordination with Increased Agent Interaction}
Though miscoordination can indeed be demonstrated in the game $R(a_x, a_y) =a_x a_y$, the problem can be mitigated through the choice of dataset. For example, by biasing the dataset to have sample means $\bar{a}_x>0$ and $\bar{a}_y>0$, something close to the optimal trajectory can be found, since the agents will move towards a high-reward region, $++$. To truly understand the severity of BRUD in offline contexts, we must look to more complex games. We present four games of increasing complexity below, showing how a higher degree of agent interaction leads to higher degrees of potential miscoordination. 

\paragraph{Decoupled Rewards: $R = a_x + a_y$.} For completeness, consider a trivial case where the shared reward yielded to agents is simply the sum of their actions. Here, $\nabla R = \nabla J = (1,1)$. Agents must simply make their actions bigger to yield higher rewards. The components of the rewards are completely decoupled, and no miscoordination occurs, regardless of the dataset used for learning.

\paragraph{Sign Agreement: $R=a_x a_y$.} As discussed before, the update in this game moves the agents in the direction of their teammate's average action in the batch, $\nabla J =( \bar{a}_y, \bar{a}_x)$. If the dataset actions happened to be biased such that $\text{sign}(\bar{a}_x)=\text{sign}(\bar{a}_y)$, then the policies will move towards a high-reward region. However, if the signs differ, the policies will move towards the low-reward region. Because there is only a single, simple interaction term, there is only a minor requirement of the dataset for successful offline learning.

\paragraph{Action Agreement: $R=-(a_x-a_y)^2$.} This game requires agents to take identical actions for optimal coordination, with anything else yielding $R<0$. The true gradient field, $\nabla R = (2a_y-2a_x,2a_x-2a_y)$, implies a line of optima, $\nabla R = 0 \iff a_x=a_y$. In contrast, under the dataset, the field is $\nabla J = (2\bar{a}_y-2a_x, 2\bar{a}_x - 2a_y)$, resulting in a single optimum, $\nabla J = 0$ at the point $( \bar{a}_y, \bar{a}_x)$. Note it is no longer the agents moving in the direction of the means of the dataset actions, but instead that the learning will \emph{converge} to this point. The requirement for optimal learning is thus no longer solely based on the signs of the mean actions, but that $\bar{x} = \bar{y}$ in the dataset, which is a strong requirement.

\begin{figure}[h]
    \centering
    \begin{subfigure}[t]{0.48\textwidth}
        \centering
        \includegraphics[width=0.9\linewidth]{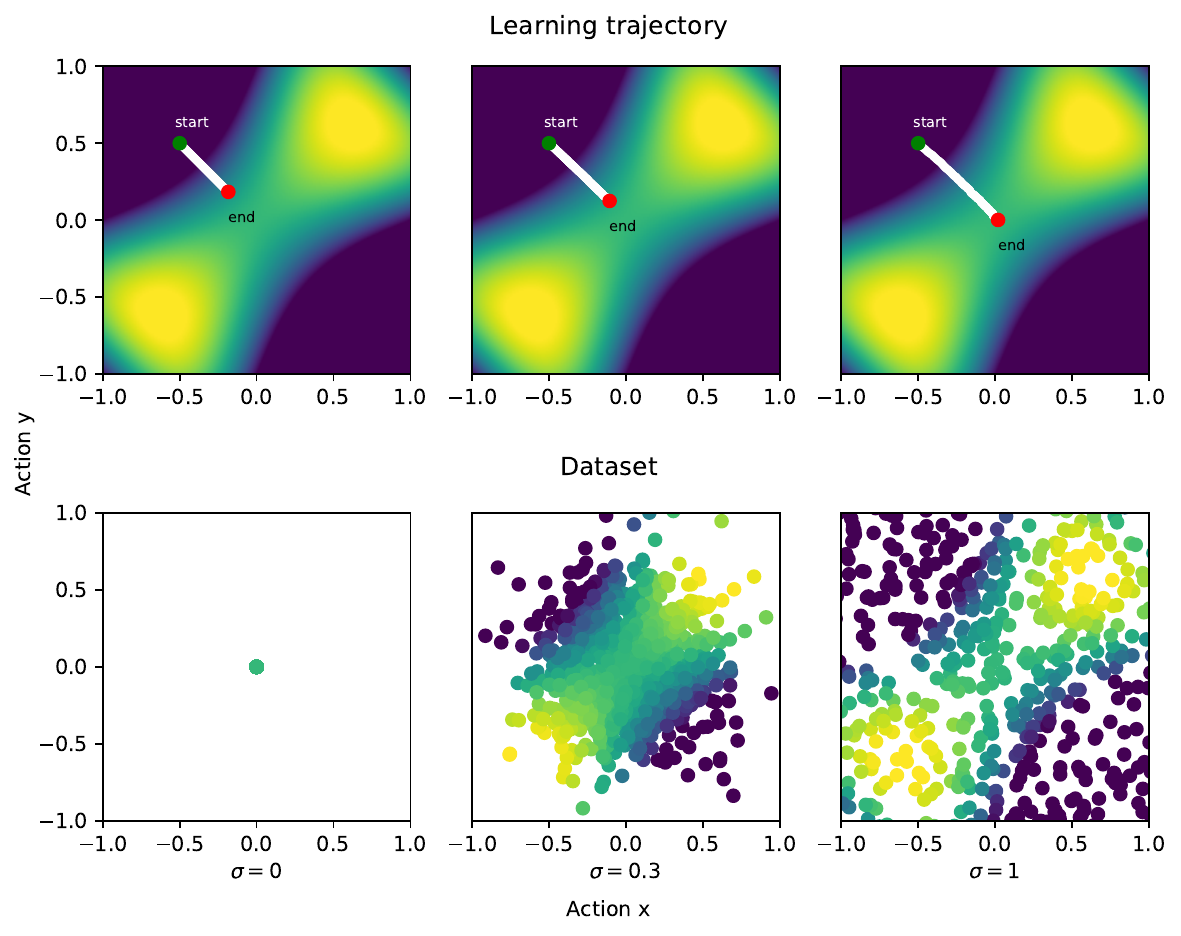}
        \caption{Origin-centred dataset, $\bar{\mathbf{a}}=\bm{0}$}
        \label{fig:twin-peaks-results-origin}
    \end{subfigure}
    \hfill
    \begin{subfigure}[t]{0.48\textwidth}
        \centering
        \includegraphics[width=0.9\linewidth]{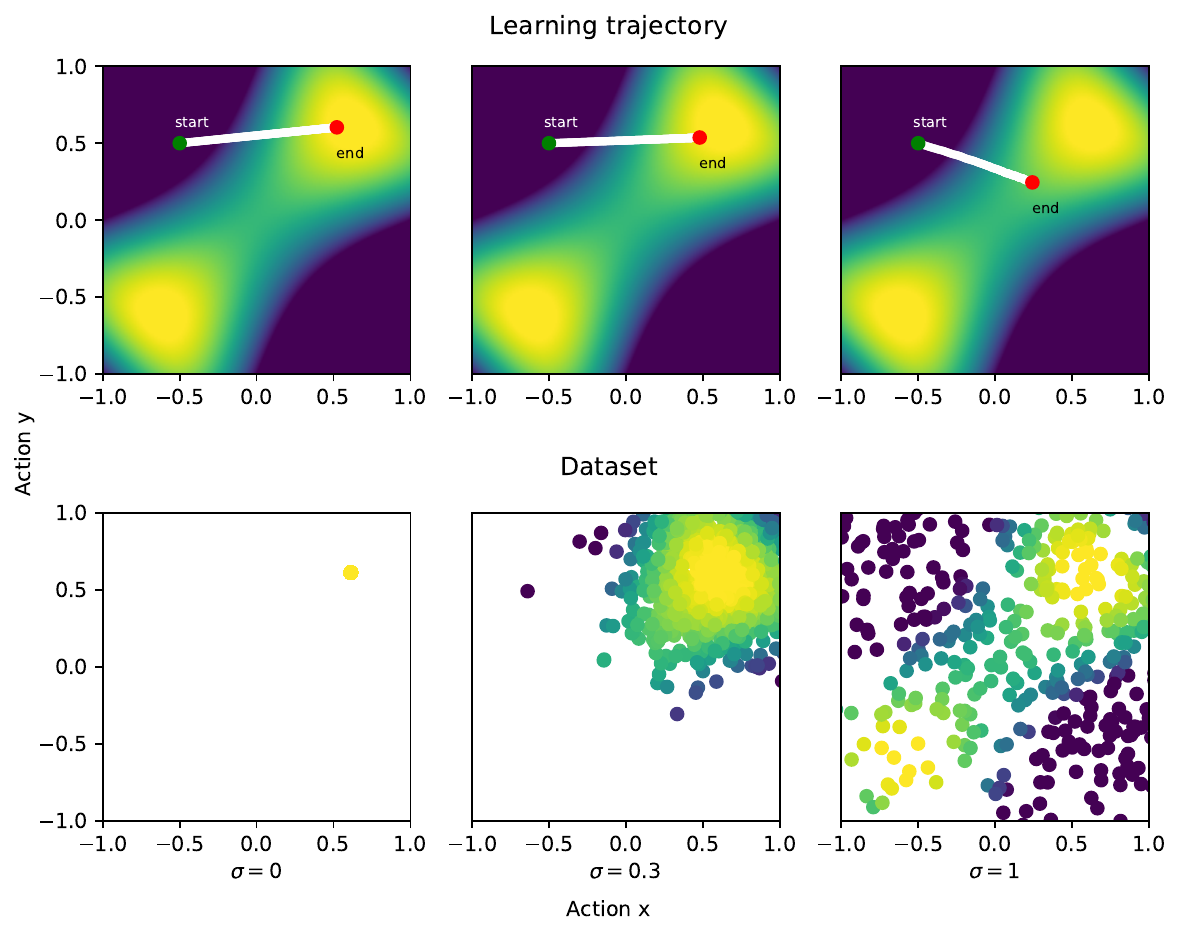}
        \caption{Optimum-centred dataset, $\bar{\mathbf{a}}=\mathbf{a}^\dagger$}
        \label{fig:twin-peaks-results-optimum}
    \end{subfigure}
    \hfill
    \caption{Visualisations from the \emph{Twin Peaks} game (with $A=1, B=4, C=5$). We see that with an origin-centred dataset (\ref{fig:twin-peaks-results-origin}), offline BRUD learning cannot find the true policy optimum, regardless of the dataset variance, always simply converging to the origin. With an optimum-centred dataset (\ref{fig:twin-peaks-results-optimum}), optimality in the learnt policy is only found if the variance is zero. As the variance increases, the learnt policy moves away from the true optimum and towards the origin. These empirical results validate the analytical solutions.}
    \label{fig:twin-peaks}
\end{figure}

\paragraph{Twin Peaks: $R=-A(a_x^2+a_y^2) - B(a_x a_y)^2 + Ca_x a_y,\; \{A>0, B>0, C>2A\}$.} Finally, we study a set of polynomial games of a higher degree, allowing for more interaction terms. For brevity, our treatment is presented for agent $x$, but all statements apply symmetrically to agent $y$, as the function itself is symmetric in the agent's actions. The surface has two peaks,
with \emph{true} maxima at $a_x^\dagger = \pm\sqrt{(C-2A)/2B}$. Most interesting in this polynomial is the bivariate quartic interaction, $(a_x a_y)^2$, since it is optimised with BRUD as $\mathbb{E}_{a_y\sim\mathcal{B}}[\nabla_{a_x} a_x^2 a_y^2] = 2a_x(\bar{a}_y^2 + \sigma_y^2)$, where $\sigma_y^2$ is the sample variance of the $y$ actions in $\mathcal{B}$. Thus, we see that the outcome of offline learning depends not only on the data's sample mean but also on its spread. Indeed, we can derive two interesting relationships when learning offline in this game. Firstly, for learning to converge to the true optimum, we have
{
\footnotesize
\begin{equation}
    \label{eqn:twin-peak_learning-to-true-optimum}
      a_x^* = a_x^\dagger \iff \sigma_y({\bar{a}_y}) = \sqrt{-{\bar{a}_y}^{\;2} \pm \left(\frac{C}{\sqrt{2B(C-2A)}}\right)\bar{a}_y-\frac{A}{B}},
\end{equation}
}
which says the dataset's standard deviation must be a function of its mean. Notice then if we centred the dataset around the origin, such that $\bar{a}_y = 0$, then $\sigma_y=\sqrt{-A/B}$, which is imaginary. Hence, there exists no distribution of data that enables learning the true optimum in this game, using offline BRUD, if the dataset is centred around $(0,0)$---even if the dataset is infinitely large. We validate this result empirically in Figure~\ref{fig:twin-peaks-results-origin}, showing that increasing the variance of the data does not help the learning succeed, for the converged policy is always simply $(0,0)$.

Secondly, the expression for the converged learnt policy is,
{
\footnotesize
\begin{equation*}
      \nabla J = 0 \iff a_x^* = \frac{C\bar{a}_y}{2A + 2B(\bar{a}_y^{\;2} + \sigma_y^2)}.
\end{equation*}
}
This expression corroborates the previous result, showing that an origin-centred dataset will always converge to the policy $a_x^* = 0$. Suppose we now centre the data exactly around the \emph{true} optimum instead, $\bar{a}_y = a_y^\dagger$. Under such conditions, learning will converge to the optimal policy only when $\sigma_y^2 = 0$ (that is, we must solely have optimal data in the dataset, with no spread); but as $\sigma_y^2 \to \infty$, then $a_x^*\to 0$, which is increasingly far away from the true optimum. This result is validated empirically in Figure~\ref{fig:twin-peaks-results-optimum}. Perhaps counter-intuitively, we thus see that increasing diversity in the dataset actions can lead to worsening performance when learning offline in this game.

\paragraph{Remark} These polynomial games indicate a clear relationship: as the degree of agent interaction increases, the requirements of the dataset become more stringent for learning to converge to the true optimum with BRUD. As a result, the possibility of miscoordination increases.

\section{Proximal Joint-Action Prioritisation for Offline Learning}
We have seen that the BRUD approach to policy learning is highly susceptible to coordination failure in the offline setting. Nonetheless, BRUD remains useful for offline learning, since it allows us to stay tightly coupled to the dataset, which is the only signal available. However, simply updating each agent's policy in response to the current joint policy learned from the data has been shown empirically to work poorly in offline MARL~\citep{cfcql}. 

Our analysis in Section 3 highlights that \textit{which} data is sampled \textit{when} could make a critical difference in the utility of best response updates. For instance, the key difference between the successful learning in Figure~\ref{fig:off-policyness}, and the coordination failure in Figure~\ref{fig:offline-uniform-data}, relates to the similarity between the current joint policy, $\bm{\theta}$, and the joint action used for the policy update, $\mathbf{a}\sim\mathcal{B}$. In fact, the problem illustrated in Figure~\ref{fig:best-response-illustration} would not have occurred if the data point, $\mathbf{a}_{(t)}$, was in the same quadrant as the current policy.

Therefore, in this work, we advocate for \emph{prioritised} dataset sampling methods as a promising area for innovation to improve learning in offline MARL. Furthermore, we consider sampling methods as an ``orthogonal'' axis to other effective approaches for offline learning such as critic and policy regularisation, where it can easily be combined with these methods to potentially great effect. As a way of demonstrating a preliminary instantiation of this idea, we propose \emph{Proximal Joint-Action Prioritisation} (PJAP) as a class of offline sampling methods. In PJAP, prioritised experience replay~\citep{per2015} is used to increase the probability of sampling actions that were generated by policies resembling the current joint policy, with priorities defined proportional to some similarity metric.

Conceptually, for each trajectory, $\tau$, in a dataset, $\mathcal{B}$, we model an underlying joint dataset-generating policy, $\bm{\beta}_\tau$. Note that a given dataset may comprise various distinct dataset-generating policies---e.g.\, when the dataset has trajectories with both low and high returns, recorded over an online training procedure. We denote the current learnt joint policy after $k$ updates as $\bm{\mu}_{(k)}$. In PJAP, we set the priority, $\rho_{k+1}$, for each of the trajectories $\tau\sim\mathcal{B}$, to be inversely proportional to some function of the distance between the current joint policy and the dataset-generating policy, $d(\bm{\mu}_{(k)}, \bm{\beta}_\tau)$. 

As a specific instance of PJAP, we propose transforming the distance on a Gaussian, $e^{-\alpha d^2}$, where $\alpha$ controls how rapidly the priorities decrease with respect to the distance. Under this transformation, we ensure small distances yield similar, large priorities, whereas larger distances yield exponentially smaller priorities. We also clip the minimum priority to some small value $\epsilon > 0$, which avoids making certain samples so unlikely that they are effectively never seen again. This parameter, $\epsilon$, is thus akin to controlling ``exploration'' of the dataset, where very occasionally we want to sample data that is, in fact, quite different to our current joint policy. In summary, our instance of PJAP takes the following prioritisation procedure,
\begin{align}
    \texttt{PJAP}_{\mathcal{N}(\epsilon)}:\quad\quad \rho_{k+1}(\tau) = \text{max}\left[e^{-\alpha d(\mu_k, \beta_\tau)^2}, \epsilon\right],
    \label{eqn:pjap}
\end{align}
We note that in practice, there are three key challenges when implementing PJAP. Firstly, we typically do not have access to the dataset-generating policy itself, $\bm{\beta}_\tau$. Thus in this work, we use the sampled actions as a proxy for the policy that generated them, and compare them to the actions taken by the agents under the sampled observations. Secondly, it is computationally unrealistic to recompute the priorities for \emph{all} trajectories in the dataset at each update step. Therefore, we fix this by bootstrapping the priority updates---updating only a subset of samples at a time. Thirdly, we concede that coming up with a good distance measure for a particular problem can be tricky, especially in higher-dimensional action spaces.

We now demonstrate our approach in the case of deterministic policies (e.g.\ MADDPG), and present different implementations of \texttt{PJAP}$_{\mathcal{N}(\epsilon)}$ using context-specific distance measures. First in the context of polynomial games, and then in the following section, a more complex MARL setting from MAMuJoCo \citep{peng2021facmac}.  We note that developing generally performant ``context-agnostic'' distance measures exists as a fruitful area for future work. 

\subsection{PJAP in Polynomial Games}
Although the datasets generated for the polynomial game are quite small, we use the L1 norm as a distance metric since ideally our chosen metric should be able to prioritise few samples from very large datasets. Another advantage of the L1 norm is that is it computationally inexpensive when compared to other distance and similarity metrics such as the L2 norm and cosine similarity. At step~$k$, with current policy parameters, $\bm{\theta}_{(k)}$, we update the priorities for each sample $\hat{\mathbf{a}} = (\hat{a}_x,\hat{a}_y)\sim\mathcal{B}$ using the PJAP formulation from Equation~\ref{eqn:pjap}, and the distance measure,
\begin{align}
d(\mu_k, \beta_\tau) \triangleq \|\hat{\mathbf{a}}, \bm{\theta}_{(k)}\|_1.
\end{align}

\begin{figure}[h]
    \centering
    \includegraphics[width=\linewidth]{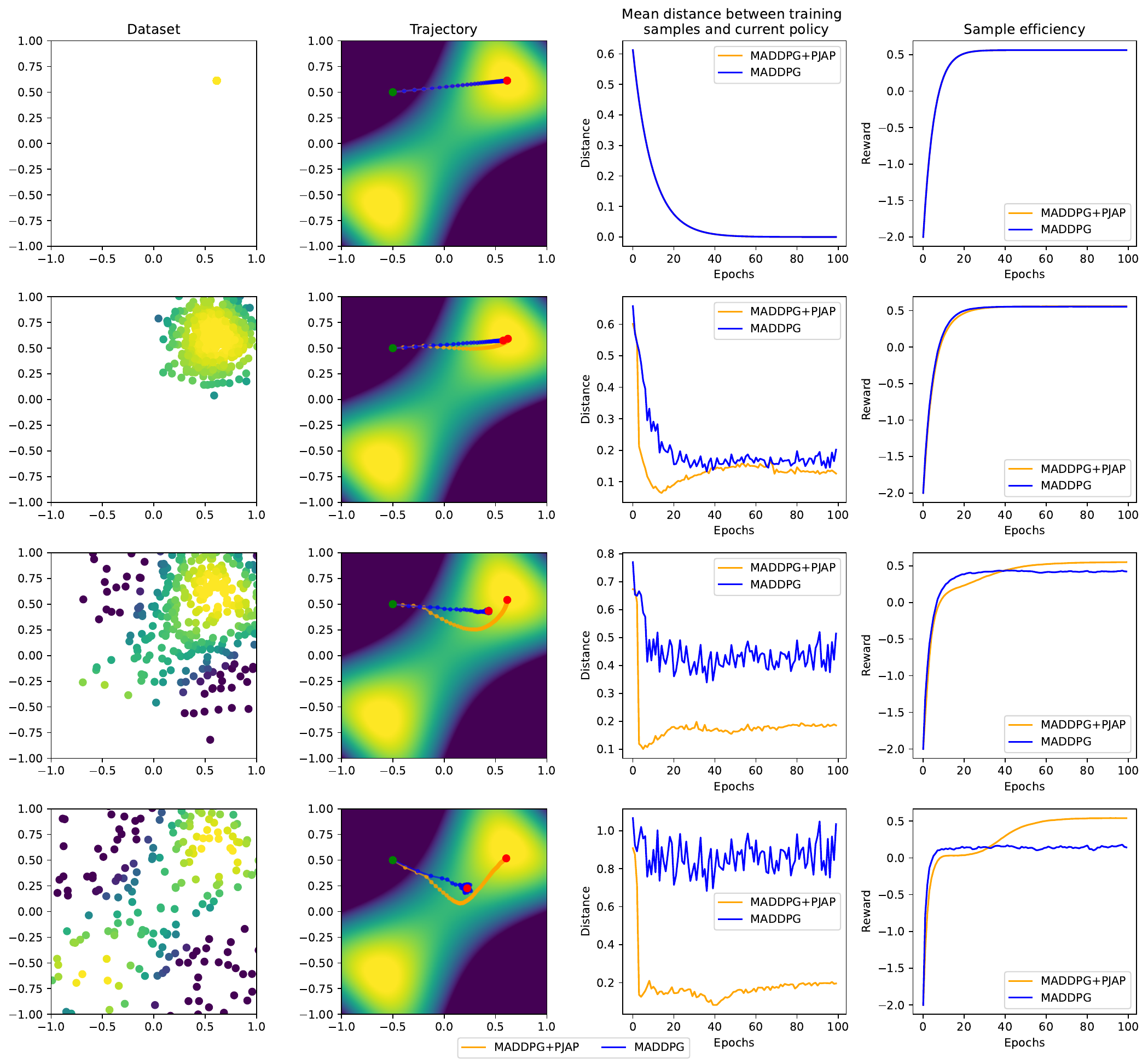}
    \caption{Results of using PJAP with MADDPG in the Twin Peaks game, fixing the problem previously seen in Figure~\ref{fig:twin-peaks-results-optimum}. Each row uses a specific dataset: all centred on the true optimum, but with increasing variances, shown in the first column. The corresponding trajectories of using MADDPG with and without PJAP are shown in the second column. The third column shows how using PJAP lowers the mean distance between sampled data and the current policy, which enables convergence to higher performance, seen in the fourth column.}
    \label{fig:twin-peaks-soln}
\end{figure}

Figure \ref{fig:twin-peaks-soln} illustrates how the failure mode seen in Figure~\ref{fig:twin-peaks-results-optimum} can be addressed by our use of PJAP. Whereas offline MADDPG using BRUD fails to converge upon the correct solution (unless $\bm\sigma=\bm{0}$), MADDPG with PJAP finds the solution consistently across the various datasets.

\subsection{PJAP in MAMuJoCo} \label{sec:halfcheetah}
\begin{figure}[h]
    \centering
    \begin{subfigure}[p]{0.6\textwidth}
        \centering
        \includegraphics[width=\linewidth]{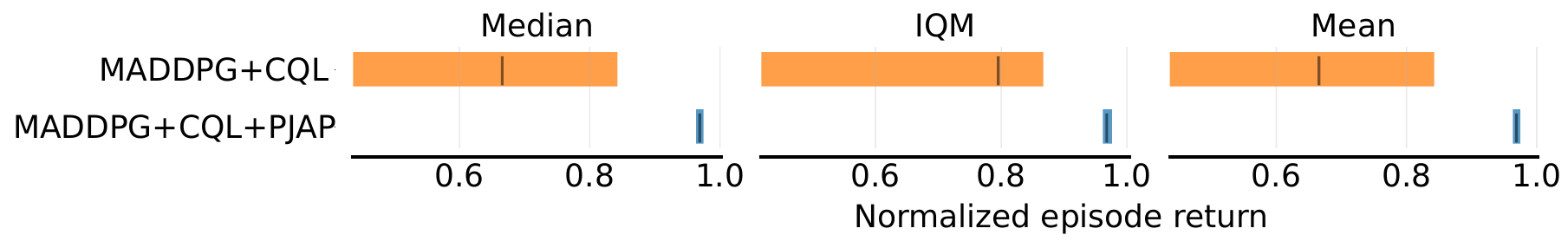}
        \caption{Performance on \texttt{Good} dataset.}
        \label{fig:mamujoco-good-final-perf}
    \end{subfigure}
    \hfill
    \begin{subfigure}[p]{0.35\textwidth}
        \centering
        \includegraphics[width=0.8\linewidth]{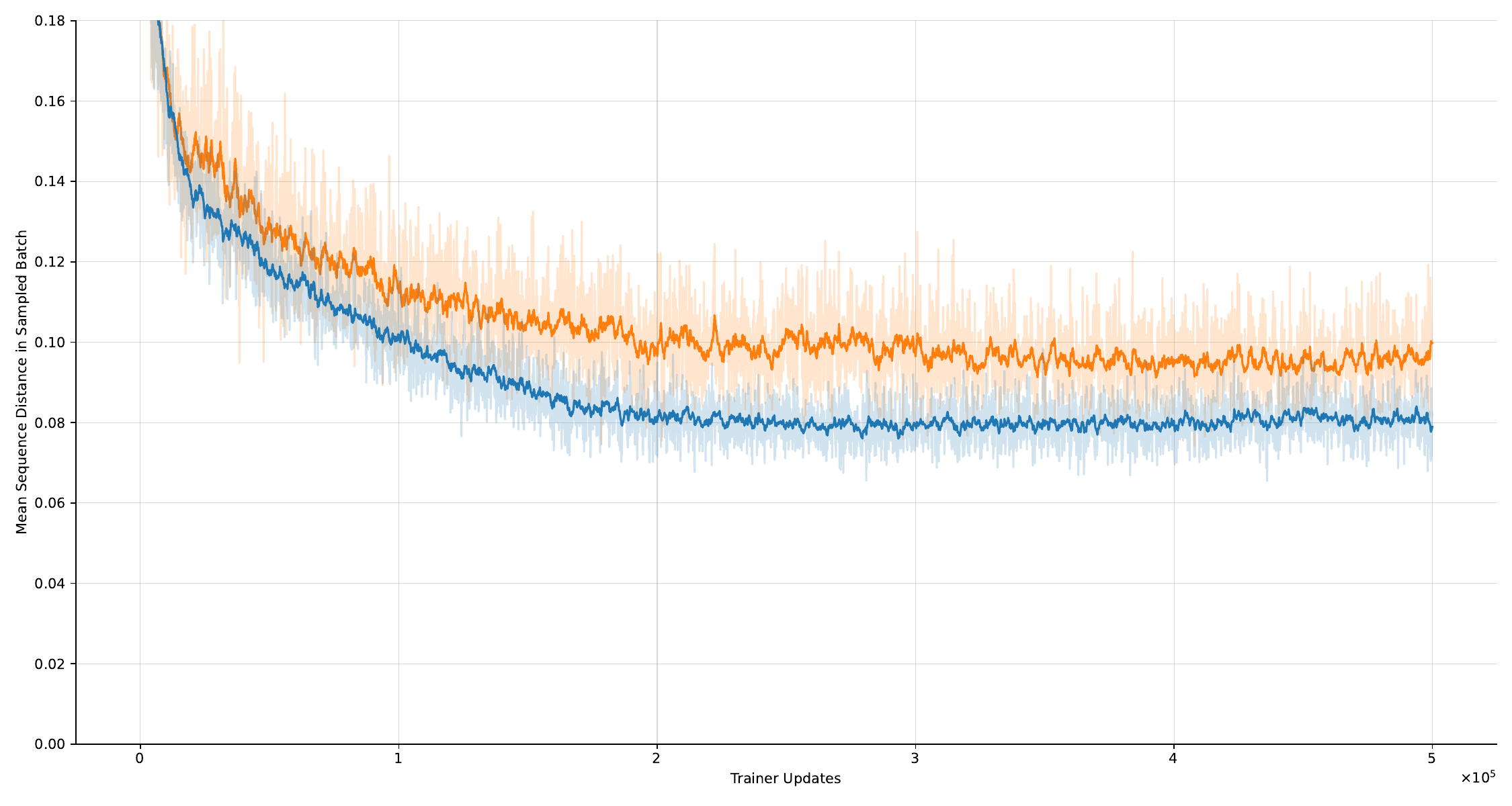}
        \caption{Mean distance plot for the \texttt{Good} dataset experiment.}
        \label{fig:mamujoco-good-distance}
    \end{subfigure}
    \vspace{1em}
    \begin{subfigure}[p]{0.6\textwidth}
        \centering
        \includegraphics[width=\linewidth]{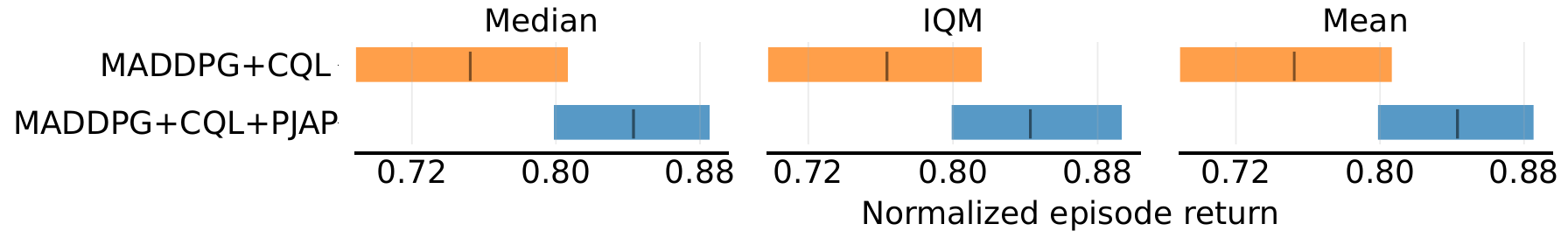}
        \caption{Performance on \texttt{Good-Medium} dataset.}
        \label{fig:fig:mamujoco-goodmedium-final-perf}
    \end{subfigure}
    \hfill
    \begin{subfigure}[p]{0.35\textwidth}
        \centering
        \includegraphics[width=0.8\linewidth]{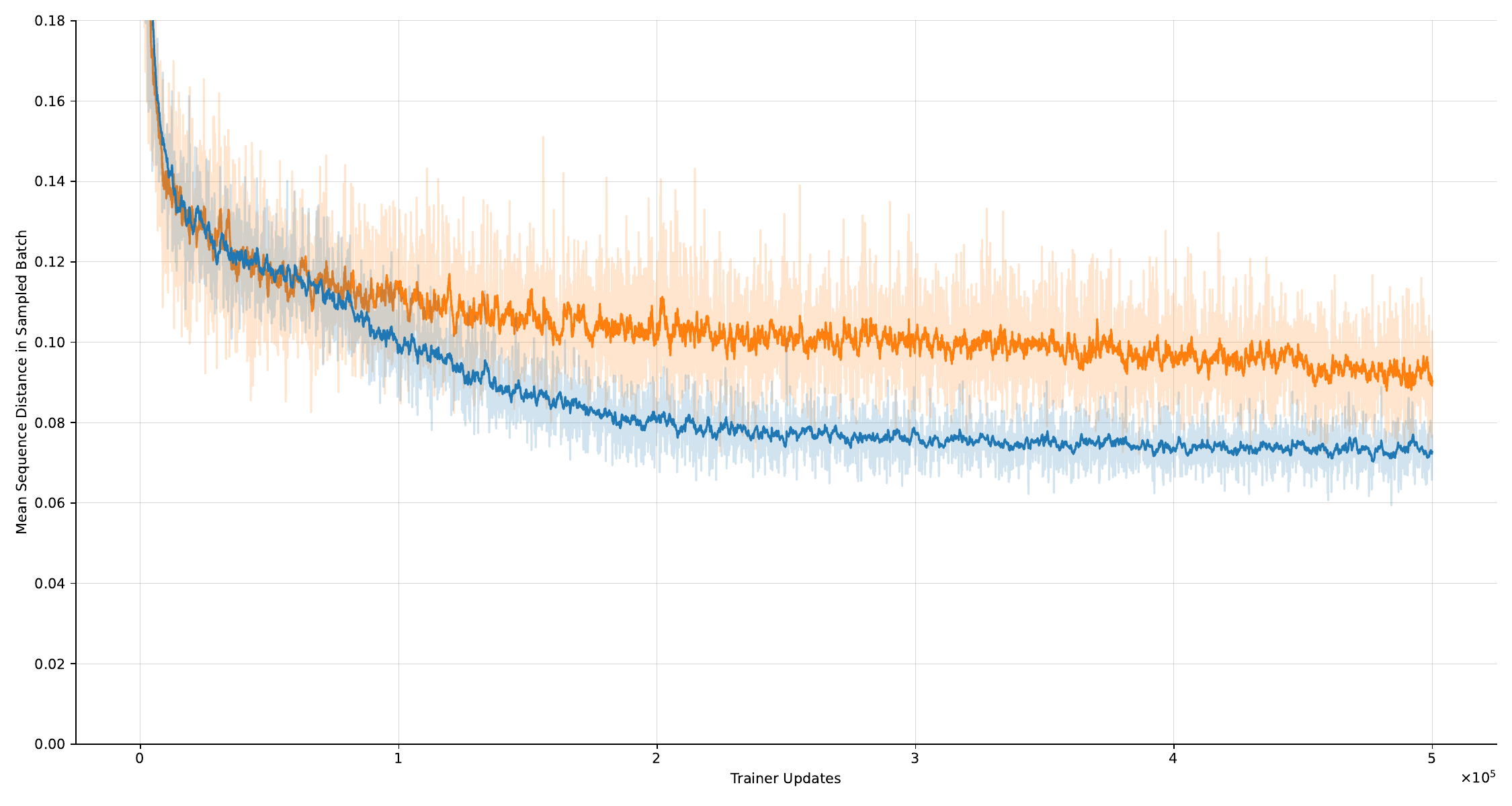}
        \caption{Mean distance plot for the \texttt{Good-Medium} dataset experiment.}
        \label{fig:mamujoco-goodmedium-distance}
    \end{subfigure}
    \caption{We compare the performance of MADDPG+CQL on 2-Agent \texttt{HalfCheetah} with and without PJAP. We experiment on the \texttt{Good} and \texttt{GoodMedium} datasets from OG-MARL~\citep{formanek2023og-marl}, showing the final performance plots (Figures \ref{fig:mamujoco-good-final-perf} and \ref{fig:fig:mamujoco-goodmedium-final-perf}) after $500k$ training steps, with bootstrap confidence intervals over 10 independent random seeds~\citep{agarwal2022deep, gorsane2022standardised}. We also show the average distance (as defined in Section \ref{sec:halfcheetah}) between actions sampled from the dataset and the current learnt policy (Figures \ref{fig:mamujoco-good-distance} and \ref{fig:mamujoco-goodmedium-distance}).}
    \label{fig:res}
\end{figure}

Next, we consider how to implement PJAP in a higher-dimensional setting, namely 2-Agent HalfCheetah from MAMuJoCo~\citep{peng2021facmac}. Here, the environment is no longer stateless and policies are conditioned on observations that change during an episode. Therefore, it is not immediately clear what a suitable distance metric can be to measure the proximity between the current learnt deterministic policy and the behaviour policies for each agent. In our experiments, we explore with using the L1 distance between the actions from the current learnt policy $\mu_k$ and the sequence of actions in a given trajectory. That is, for a given trajectory $\tau\sim\mathcal{D}$, we consider the sequence of observations and actions which make up the trajectory $\tau=\{((o_t^1,\dots,o_t^n), (a_t^1,\dots,a_t^n)),\dots, ((o_T^1,\dots,o_T^n), (a_T^1,\dots,a_T^n)) \}$. Each $a_t^i$ in the trajectory $\tau$ comes from the unknown behaviour policy $\beta_\tau$. Thus, the average L1 distance between action $a_t^i$ and $\mu^i_k(o^i_k)$ can be seen as an approximate distance metric for PJAP.
{
\footnotesize
\begin{align} 
    d(\mu_k, \beta_\tau) \triangleq \frac{1}{N\cdot T} \sum_i^N\sum_t^T \|\mu(o_{i(t)}), a_{i(t)}\|_1.
    \label{eq:distance_halfcheetah}
\end{align}
}
We use the MADDPG+CQL implementation from OG-MARL \citep{formanek2023og-marl} as our baseline and then compare it to a version of MADDPG+CQL where we incorporate PJAP using the distance metric in Equation \ref{eq:distance_halfcheetah} and the prioritisation function in Equation \ref{eqn:pjap}. We evaluate our method against the baseline on two different datasets from the \texttt{2-Agent HalfCheetah} environment. The results from the experiments are given in Figure \ref{fig:res}. The average distance between actions sampled from the dataset and the current learnt policy is also given to highlight that PJAP reduces this distance as compared to the baseline, which corroborates the findings from the polynomial game experiment. All hyperparameters between the baseline and our variant with PJAP are kept constant and have been included in the attached code.

\section{Related Work}
Notable progress has been made in the field of offline MARL in recent years. \cite{jiang2023offline} highlighted that in offline MARL the transition dynamics in the dataset can significantly differ from those of the learned policies, leading to coordination failures. They address this by normalising the transition probabilities in the dataset. \cite{yang2021believe} highlight and address the rapid accumulation of extrapolation error due to \textit{Out-of-Distribution} actions in Offline MARL. \cite{omar} show that MADDPG and Independent DDPG policy gradients struggle to optimise conservative Q-functions \citep{kumar2020cql}, and propose using a zeroth-order optimisation to learn more coordinated behaviour. \citet{cfcql} also highlight the limitations of MADDPG+CQL and propose a per-agent CQL regularisation that scales better in the number of agents. \cite{omiga} also explored a novel approach to regularising the value function in Offline MARL. \citet{tian2023goodtrajectories} consider an imbalance of agent expertise in a dataset, which can contaminate the offline learning of all agents. They address this problem by learning decomposed rewards, and then reconstructing the dataset while favouring high-return individual trajectories. Finally, work by \citet{cui2022offline} addresses the additional fundamental difficulties of solving multi-agent learning problems using a static dataset. The authors show that the dataset requirements for the solvability of a two-player zero-sum Markov game are stricter than for a single-strategy setting.

\section{Discussion} \label{sec:conclusion}
In this paper, we use simple two-player polynomial games to highlight and study the fundamental problem of miscoordination in offline MARL, when using a Best Response Under Data (BRUD) approach to policy learning. Building on our analyses, we propose Proximal Joint Action Prioritisation (PJAP), where sampled experience data is prioritised as a function of the current joint policy being learnt. We instantiate an instance of PJAP, and demonstrate how it can solve miscoordination problems in both the simplified polynomial game case, and in the more complex MARL setting of MAMuJoco.

Importantly, though, our work primarily aims to be a catalyst for further development of dataset sampling prioritisation methods, as one tool in our offline MARL toolkit. This tool exists alongside other offline MARL remedies, such as critic and policy regularisation, all helping mitigate the difficulties of offline learning, together. We believe that PJAP paves the way for interesting research ideas in offline MARL.

\paragraph{Limitations}
\label{sect:limitations}
This paper primarily focuses on theoretical contributions and insights in simplified settings, using polynomial games as a backbone. Though useful as an interpretable and accessible tool, the context is admittedly limited in several ways---it is stateless, comprises just two agents and assumes perfect knowledge of the reward surface. We acknowledge that these limitations constrain the generality of our conclusions, even when supported with more complex empirical results. However, remain confident that our approach takes an important step to improving our understanding of coordination in offline MARL.

\newpage
\bibliographystyle{abbrvnat}
\bibliography{references}

\begin{thebibliography}{23}
\providecommand{\natexlab}[1]{#1}
\providecommand{\url}[1]{\texttt{#1}}
\expandafter\ifx\csname urlstyle\endcsname\relax
  \providecommand{\doi}[1]{doi: #1}\else
  \providecommand{\doi}{doi: \begingroup \urlstyle{rm}\Url}\fi

\bibitem[Agarwal et~al.(2022)Agarwal, Schwarzer, Castro, Courville, and Bellemare]{agarwal2022deep}
R.~Agarwal, M.~Schwarzer, P.~S. Castro, A.~Courville, and M.~G. Bellemare.
\newblock Deep reinforcement learning at the edge of the statistical precipice, 2022.

\bibitem[Barde et~al.(2024)Barde, Foerster, Nowrouzezahrai, and Zhang]{barde2023modelbased}
P.~Barde, J.~Foerster, D.~Nowrouzezahrai, and A.~Zhang.
\newblock A model-based solution to the offline multi-agent reinforcement learning coordination problem.
\newblock \emph{Proceedings of the 23rd International Conference on Autonomous Agents and Multiagent Systems}, 2024.

\bibitem[Cui and Du(2022)]{cui2022offline}
Q.~Cui and S.~S. Du.
\newblock When are offline two-player zero-sum markov games solvable?
\newblock \emph{Advances in Neural Information Processing Systems}, 35:\penalty0 25779--25791, 2022.

\bibitem[Dresher et~al.(1950)Dresher, Karlin, and Shapley]{dresher1950polynomial}
M.~Dresher, S.~Karlin, and L.~S. Shapley.
\newblock Polynomial games.
\newblock \emph{Contributions to the Theory of Games I}, 1950.

\bibitem[Formanek et~al.(2023)Formanek, Jeewa, Shock, and Pretorius]{formanek2023og-marl}
C.~Formanek, A.~Jeewa, J.~Shock, and A.~Pretorius.
\newblock Off-the-grid marl: Datasets and baselines for offline multi-agent reinforcement learning.
\newblock In \emph{Proceedings of the 2023 International Conference on Autonomous Agents and Multiagent Systems}, AAMAS '23, page 2442–2444, Richland, SC, 2023. International Foundation for Autonomous Agents and Multiagent Systems.
\newblock ISBN 9781450394321.

\bibitem[Gorsane et~al.(2022)Gorsane, Mahjoub, de~Kock, Dubb, Singh, and Pretorius]{gorsane2022standardised}
R.~Gorsane, O.~Mahjoub, R.~de~Kock, R.~Dubb, S.~Singh, and A.~Pretorius.
\newblock Towards a standardised performance evaluation protocol for cooperative marl, 2022.

\bibitem[Jiang and Lu(2021)]{jiang2023offline}
J.~Jiang and Z.~Lu.
\newblock Offline decentralized multi-agent reinforcement learning, 2021.

\bibitem[Kumar et~al.(2020)Kumar, Zhou, Tucker, and Levine]{kumar2020cql}
A.~Kumar, A.~Zhou, G.~Tucker, and S.~Levine.
\newblock Conservative q-learning for offline reinforcement learning.
\newblock \emph{Advances in Neural Information Processing Systems}, 2020.

\bibitem[Lowe et~al.(2017)Lowe, Wu, Tamar, Harb, Abbeel, and Mordatch]{lowe2017maddpg}
R.~Lowe, Y.~I. Wu, A.~Tamar, J.~Harb, P.~Abbeel, and I.~Mordatch.
\newblock Multi-agent actor-critic for mixed cooperative-competitive environments.
\newblock \emph{Advances in neural information processing systems}, 2017.

\bibitem[Mnih et~al.(2013)Mnih, Kavukcuoglu, Silver, Graves, Antonoglou, Wierstra, and Riedmiller]{dqn2013}
V.~Mnih, K.~Kavukcuoglu, D.~Silver, A.~Graves, I.~Antonoglou, D.~Wierstra, and M.~Riedmiller.
\newblock Playing atari with deep reinforcement learning.
\newblock \emph{NIPS Deep Learning Workshop}, 2013.

\bibitem[Pan et~al.(2022)Pan, Huang, Ma, and Xu]{omar}
L.~Pan, L.~Huang, T.~Ma, and H.~Xu.
\newblock Plan better amid conservatism: Offline multi-agent reinforcement learning with actor rectification.
\newblock In \emph{International conference on machine learning}, pages 17221--17237. PMLR, 2022.

\bibitem[Papoudakis et~al.(2021)Papoudakis, Christianos, Sch{\"a}fer, and Albrecht]{papoudakisBenchmarkingMultiAgentDeep2021}
G.~Papoudakis, F.~Christianos, L.~Sch{\"a}fer, and S.~V. Albrecht.
\newblock Benchmarking {{Multi-Agent Deep Reinforcement Learning Algorithms}} in {{Cooperative Tasks}}.
\newblock \emph{Proceedings of the Neural Information Processing Systems Track on Datasets and Benchmarks}, 2021.

\bibitem[Peng et~al.(2021)Peng, Rashid, de~Witt, Kamienny, Torr, Böhmer, and Whiteson]{peng2021facmac}
B.~Peng, T.~Rashid, C.~A.~S. de~Witt, P.-A. Kamienny, P.~H.~S. Torr, W.~Böhmer, and S.~Whiteson.
\newblock Facmac: Factored multi-agent centralised policy gradients, 2021.

\bibitem[Prudencio et~al.(2023)Prudencio, Maximo, and Colombini]{prudencio2023offlinerl}
R.~F. Prudencio, M.~R. Maximo, and E.~L. Colombini.
\newblock A survey on offline reinforcement learning: Taxonomy, review, and open problems.
\newblock \emph{IEEE Transactions on Neural Networks and Learning Systems}, 2023.

\bibitem[Pu et~al.(2021)Pu, Wang, Yang, Yao, and Li]{Pu2021DecomposedSA}
Y.~Pu, S.~Wang, R.~Yang, X.~Yao, and B.~Li.
\newblock Decomposed soft actor-critic method for cooperative multi-agent reinforcement learning.
\newblock \emph{ArXiv}, abs/2104.06655, 2021.

\bibitem[Rashid et~al.(2020)Rashid, Samvelyan, De~Witt, Farquhar, Foerster, and Whiteson]{rashid2020monotonic}
T.~Rashid, M.~Samvelyan, C.~S. De~Witt, G.~Farquhar, J.~Foerster, and S.~Whiteson.
\newblock Monotonic value function factorisation for deep multi-agent reinforcement learning.
\newblock \emph{Journal of Machine Learning Research}, 21\penalty0 (178):\penalty0 1--51, 2020.

\bibitem[Schaul et~al.(2016)Schaul, Quan, Antonoglou, and Silver]{per2015}
T.~Schaul, J.~Quan, I.~Antonoglou, and D.~Silver.
\newblock Prioritized experience replay.
\newblock In \emph{International Conference on Learning Representations}, 2016.

\bibitem[Shao et~al.(2023)Shao, Qu, Chen, Zhang, and Ji]{cfcql}
J.~Shao, Y.~Qu, C.~Chen, H.~Zhang, and X.~Ji.
\newblock Counterfactual conservative q learning for offline multi-agent reinforcement learning.
\newblock \emph{Advances in Neural Information Processing Systems}, 37, 2023.

\bibitem[Tian et~al.(2023)Tian, Kuang, Liu, and Wang]{tian2023goodtrajectories}
Q.~Tian, K.~Kuang, F.~Liu, and B.~Wang.
\newblock Learning from good trajectories in offline multi-agent reinforcement learning.
\newblock \emph{Proceedings of the AAAI Conference on Artificial Intelligence}, 37:\penalty0 11672--11680, 06 2023.
\newblock \doi{10.1609/aaai.v37i10.26379}.

\bibitem[Wang et~al.(2023)Wang, Xu, Zheng, and Zhan]{omiga}
X.~Wang, H.~Xu, Y.~Zheng, and X.~Zhan.
\newblock Offline multi-agent reinforcement learning with implicit global-to-local value regularization.
\newblock \emph{Advances in Neural Information Processing Systems}, 37, 2023.

\bibitem[Williams(1992)]{williamsSimpleStatisticalGradientfollowing}
R.~J. Williams.
\newblock Simple statistical gradient-following algorithms for connectionist reinforcement learning.
\newblock \emph{Machine Learning}, 8\penalty0 (3):\penalty0 229--256, May 1992.
\newblock ISSN 1573-0565.
\newblock \doi{10.1007/BF00992696}.

\bibitem[Yang et~al.(2021)Yang, Ma, Li, Zheng, Zhang, Huang, Yang, and Zhao]{yang2021believe}
Y.~Yang, X.~Ma, C.~Li, Z.~Zheng, Q.~Zhang, G.~Huang, J.~Yang, and Q.~Zhao.
\newblock Believe what you see: Implicit constraint approach for offline multi-agent reinforcement learning, 2021.

\bibitem[Zhong et~al.(2024)Zhong, Kuba, Feng, Hu, Ji, and Yang]{zhong2024heterogeneous}
Y.~Zhong, J.~G. Kuba, X.~Feng, S.~Hu, J.~Ji, and Y.~Yang.
\newblock Heterogeneous-agent reinforcement learning.
\newblock \emph{Journal of Machine Learning Research}, 25:\penalty0 1--67, 2024.

\end{thebibliography}

\newpage

\end{document}